\documentclass[sigplan,screen]{acmart}

\makeatletter
\renewcommand\@formatdoi[1]{\ignorespaces}
\makeatother

\makeatletter
\def\@copyrightspace{\relax}
\makeatother

\AtBeginDocument{%
  }

\usepackage{hyperref}
\usepackage{hyperxmp}
\usepackage{tikz}
\usepackage{amsmath}
\usepackage{wrapfig}

\usepackage[english]{babel}
\usepackage{blindtext}
\usepackage{lipsum}
\usepackage{amsthm}
\usepackage[ruled,vlined,linesnumbered]{algorithm2e}

\usepackage{filecontents}
\usepackage{xspace}
\usepackage{xcolor} 
\usepackage{hhline}
\usepackage{multirow}
\usepackage{soul}
\usepackage{subcaption}

\usepackage{array}
\usepackage{mathtools}
\usepackage{comment}

\usepackage{pifont}
\usepackage{marvosym}

\usepackage{csquotes}
\usepackage[font=itshape]{quoting}
\usepackage{colortbl}

\newcommand{\cameraold}[1]{{\color{red}{}}}

\newcommand{\jc}[1]{{\color{red}{\footnotesize [JC: #1]\xspace}}}
\newcommand{\yh}[1]{{\color{red}{\footnotesize [YH: #1]\xspace}}}

\newcommand{\ignore}[1]{{\xspace}}
\newcommand{\REMOVE}[1]{}
\newcommand{\tempstrike}[1]{}

\newcounter{packednmbr}

\newenvironment{packeditemize}{\begin{list}{$\bullet$}{
\setlength{\itemsep}{0.5pt}\addtolength{\labelwidth}{-4pt}\setlength{\leftmargin}{2.5ex}\setlength{\listparindent}{\parindent}\setlength{\parsep}{1pt}\setlength{\topsep}{2pt}}}{\end{list}}

\newcommand{\tightcaption}[1]{\vspace{-0.4cm}\caption{{\normalfont{\textit{{#1}}}}}\vspace{-0.4cm}}

\vspace{-0.05cm}

\newcommand{\eg}{{\it e.g.,}\xspace}
\newcommand{\ie}{{\it i.e.,}\xspace}

\newcommand{\mypara}[1]{\vspace{0.1cm}\noindent{\bf {#1}:}~}

\newcommand{\myparaq}[1]{\vspace{0.1cm}\noindent{\bf {#1}?}~}

\definecolor{LightCyan}{rgb}{0.88,1,1}

\setcopyright{none}
\acmISBN{}
\acmYear{}
\acmDOI{}

\acmConference[HotInfra '24]{The 2nd Workshop on Hot Topics in System Infrastructure (HotInfra'24), co-located with SOSP'24}{November 3,
  2024}{Austin, TX, USA}

\begin{document}

\date{}



\title{\fontsize{16}{10}\selectfont Do Large Language Models Need a Content Delivery Network?}


\renewcommand{\shorttitle}{Title}
\author{
Yihua Cheng, \xspace\xspace Kuntai Du, \xspace\xspace Jiayi Yao, \xspace\xspace Junchen Jiang \\ 
\textit{The University of Chicago}
}

\pagestyle{empty}
\thispagestyle{empty}

\begin{abstract}

As the use of large language models (LLMs) expands rapidly, so does the range of knowledge needed to supplement various LLM queries.
Thus, enabling modular and efficient injection of new knowledge in LLM inference is critical.
We argue that compared to the more popular fine-tuning and in-context learning, using KV caches as the medium of knowledge could simultaneously improve the modularity of knowledge injection and the efficiency of LLM serving with low cost and fast response. 
To make it practical, we envision a {\em Knowledge Delivery Network} (KDN), a new component in LLM services that dynamically optimizes the storage, transfer, and composition of KV cache across LLM engines and other compute and storage resources. 
Just like content delivery networks (CDNs), such as Akamai, enabled the success of the Internet ecosystem through their efficient data delivery, KDNs will be critical to the success of LLM applications through their efficient knowledge delivery. 
An open-sourced KDN prototype: \href{https://github.com/LMCache/LMCache}{\textbf{https://github.com/LMCache/LMCache}}. 

\end{abstract}

\maketitle

\vspace{-4pt}
\section{Background and Motivation}


Traditionally, machine learning models, such as computer vision~\cite{resnet,efficientdet,llava,ssd} and image generation~\cite{gan,stable-diffusion,sora}, learn all the knowledge from the training data.
However, with the rapid usage growth of large language models (LLMs),
applications require the ability to inject external knowledge, unseen in training, into LLM inference.
For instance, chatbots~\cite{chatbot-arena,chatgpt,characterai} and agent systems use the chatting histories as supplementary knowledge to generate personalized responses; 
enterprises use LLM agents to answer queries based on their internal databases, as new knowledge, using retrieval-augmented generation (RAG)~\cite{rag1,rag2,rag3,factual-llm1,factual-llm2}; 
and searching engines use LLM to read fresh and relevant web data from the internet for each user query~\cite{perplexity,google-enterprise,concensus}.

Not only is providing more external knowledge among the most efficient ways to boost LLM quality~\cite{yue2024inference,factual-llm1,factual-llm2,incontext1}, but the requirements of knowledge injection also become more complex and diverse. 
For instance, in enterprise settings, data scientists and application operators often demand the flexibility that allows them to directly specify which documents (or which parts of a document) should or should {\em not} be used as the context to answer a given query~\cite{Instruct69, convivablog}. 
Moreover, as new data is being constantly collected~\cite{websiteupdate, perplexityblog1}, the knowledge-injection method must be efficient enough to keep up with the rapidly evolving pool of knowledge. 

Thus, the key question is {\em ``how to design and implement a system to inject knowledge to LLMs?''} 

Three knowledge-injection methods exist. 
{\em In-context} learning and {\em fine-tuning} are the two popular options employed by knowledge-augmented generation. 
Fine-tuning retrains an LLM on a new dataset using a set of labeled knowledge to update all or parts of the model's parameters.
Unlike fine-tuning, in-context learning leverages the existing capabilities of the model to interpret and respond to new information, {\em without} altering the model's parameters. 
Yet, compared to using a fine-tuning model, in-context learning has much higher run-time computation overhead as it has to process a much longer input (\ie the prefill step), before the model can generate the first token.

An alternative, which we refer to as {\em KV-cache learning}, is to let the LLM pre-compute the {\em KV cache}\footnote{KV cache is the intermediate state when LLM prefill on a text~\cite{cachegen,yao2024cacheblend,scissorhands}, which represents the LLM's understanding of the knowledge.} of the new-knowledge text so that when the same knowledge is needed to supplement another LLM query, the KV cache can be directly used by LLM. 
This way, LLMs can directly generate the response as fast as the fine-tuned model, without the extra computation overhead to prefill the text of the knowledge as in-context learning. 
However, KV-cache learning, with a straightforward implementation, will suffer from the large size of the KV caches. 

Most research in the machine-learning communities has primarily focused on the generation quality of different knowledge-injection methods, in F1 score and accuracy~\cite{fine-vs-incontext1,fine-vs-incontext2,fine-vs-incontext3}. 
Studies have shown that both in-context learning and fine-tuning can achieve high text-generation quality if they are configured appropriately by {\em machine-learning} engineers~\cite{alghisi2024should, zhang2024raft, gupta2024rag}.
However, less is known about the tradeoffs of these methods presented to {\em system engineers} who implement and maintain the LLM infrastructure for knowledge-augmented LLM applications.

\begin{figure}[t!]
    \centering
    \includegraphics[width=0.95\columnwidth]{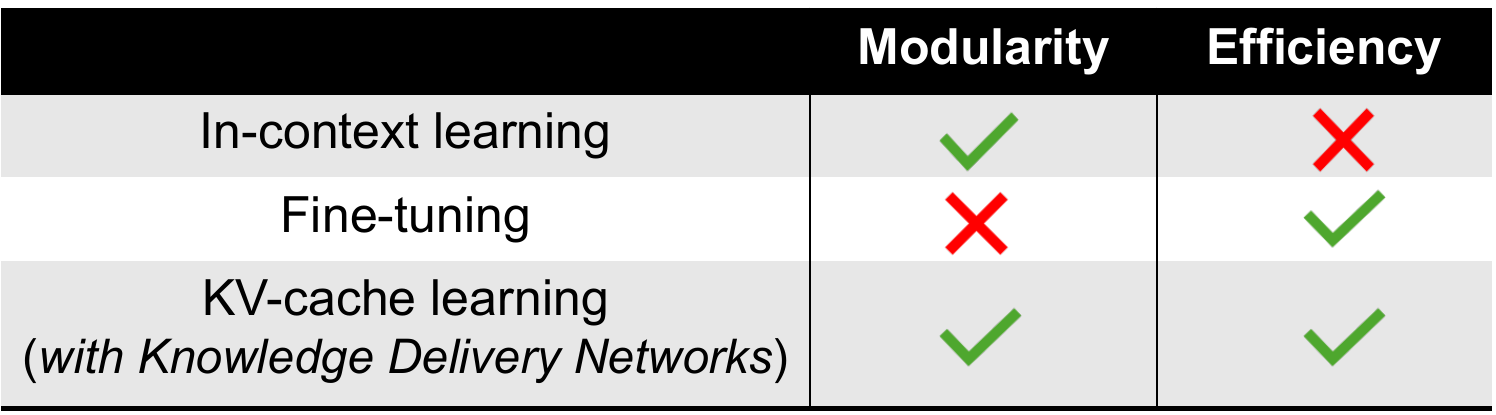}
    \tightcaption{Knowledge-injection methods make trade-offs between {\bf modularity} and {\bf efficiency}. With Knowledge Delivery Networks (KDNs), KV-cache learning can improve on both dimensions compared to in-context learning and fine-tuning.
    }
    \label{fig:comparison}
\end{figure}

This paper sheds light on these methods from a {\bf \em system architecture's}\footnote{Architecture refers to the interface between the modules, and we leave the refinement of implementation to future work.} perspective (Figure~\ref{fig:comparison}). 
Specifically, we make two key arguments:

\begin{packeditemize}
\item {\em Existing knowledge-injection methods---in-context learning, fine-tuning, and KV-cache learning---mainly differ in the tradeoffs between their {\bf modularity} (ease of adding new knowledge and flexibility to specify injected knowledge) and {\bf efficiency} (in per-query cost and response delay).} (\S\ref{sec:tradeoffs})

\item {\em Compared to in-context learning and fine-tuning, KV-cache learning {\bf could} improve both modularity and efficiency, if a new system component, called {\bf knowledge-delivery network (KDN)}, optimizes the management of KV caches by harnessing several emerging research efforts.} (\S\ref{sec:kdn})

\end{packeditemize}

At a high level, a KDN serves as a backend of LLM-processed knowledge (\ie KV caches), which can be a part of an LLM-serving system or shared across multiple LLM-serving systems. 
Unlike the existing LLM-serving systems~\cite{kwon2023vllm, huggingface_tgi_2024, sglang}, which deeply couple KV caches with LLM engines (\ie binding KV caches with GPUs and managing KV caches inside inferencing), KDNs call for a clean {\bf\em separation} between KV-cache management and LLM serving engines, for better modularity and efficiency. 

We will outline the key components of a KDN, including a storage pool of KV caches that leverages KV-cache compression, a fast KV-cache streaming system to transfer KV caches between LLM serving engines, and a KV-cache blender module that dynamically puts together multiple pieces of knowledge stored in modularized KV caches. 
Using a prototype of KDN, we will show that some emerging research efforts have already provided preliminary techniques, which together can make highly efficient KDNs a reality.

\vspace{-2pt}
\section{LLM Knowledge-Injection From A System Perspective}
\label{sec:tradeoffs}

\begin{figure*}
    \centering
    \includegraphics[width=0.99\linewidth]{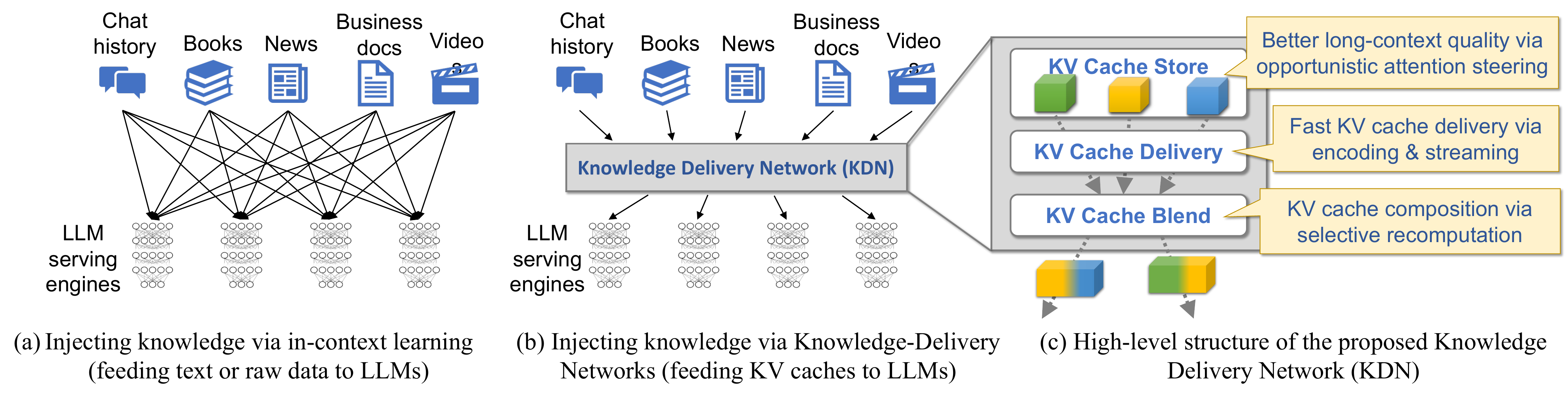}
    \tightcaption{Architecture of a Knowledge Delivery Network (KDN).}
    \label{fig:kdn}
\end{figure*}

Knowledge-augmented generation, particularly, fine-tuning and in-context learning, is well-studied in the AI literature. 

{\em Fine-tuning} embeds a corpus of texts in the LLM's weights, so the fine-tuned model can directly respond to a user query with a low response delay. 
However, as the entire corpus of texts must be embedded in the model together, fine-tuning lacks 
the flexibility to add new knowledge and specify what knowledge should or should not be used.

{\em In-context learning} is the opposite of fine-tuning, as it allows the operators to specify which external knowledge should be used easily by putting the texts to the LLM's input.
However, the compute overhead of prefilling will grow superlinearly with the input length, causing a long response delay when more external data is added to the input.

\vspace{-2pt}
\subsection{The Efficiency-Modularity Tradeoff}

Both methods can achieve similar text-generation quality if used in the right way~\cite{gupta2024rag, zhang2024raft, alghisi2024should}.
Instead of viewing these options only in terms of accuracy (\ie through the ML perspective), 
we compare the knowledge-injection methods along the following two {\em system-centric} metrics: {\bf modularity} and {\bf efficiency}.

\mypara{Modularity}
In the context of knowledge-augmented LLM, the modularity of a method includes two aspects. 
First, a modular approach should allow service providers to specify which knowledge to use and compose them easily.
Second, the overhead (\eg time, cost) of injecting new knowledge into the LLM should be minimized. 

In-context learning puts the new knowledge in the model's input, rather than the model itself. 
The separation of knowledge and model serves as the key to modularity---LLM service providers can specify which knowledge to use and easily compose different pieces of knowledge, which helps the LLM to avoid conflicting knowledge and improve the generation quality.
In contrast, fine-tuning has poor modularity.
Users cannot specify which knowledge in the fine-tuned model would be used to generate the answer.
Moreover, fine-tuning needs to happen for every new knowledge and model, which may take hours to days to complete.

\mypara{Efficiency}
On the other hand, the efficiency of a knowledge-augmented LLM system is measured by per-query cost and response delay during LLM inference.
Cost is the computation used for the LLM to handle a request, and response delay is defined as the time between the LLM receiving the request and the generation of the first token.
Viewed through this lens, in-context learning is not ideal, because when using in-context learning, LLMs must spend a long time {\em prefilling} input text with the knowledge before generating the first token.
In contrast, fine-tuning is better in terms of efficiency.
Because the knowledge is embedded in the model's weights, the fine-tuned model can skip the long prefill.

In short, in-context learning is more modular but sacrifices efficiency, while fine-tuning, though achieving better efficiency, suffers from the overhead of incorporating and controlling the knowledge due to poor modularity.

\vspace{-16pt}
\subsection{KV-Cache Learning}

Alternatively, {\em KV cache learning} stores the knowledge in LLM-generated KV cache, and injects knowledge by feeding the KV cache to the LLM, without modifying the model. 
A KV cache stores the knowledge in the form of the attention state generated by the model after it processes the text, so the KV cache, once generated, can be reused by the LLM to skip prefilling if the subsequent requests use the same context.
When many queries reuse the same context, 
reusing its KV cache {\em could} reduce the per-query delay and compute usage, while still preserving the modularity as in-context learning.
The idea of {\em KV-cache learning} has gained increasing attention in LLM services (\eg~\cite{sglang,lmcache-release,qin2024mooncake}). 

\myparaq{Why storing knowledge in KV caches pays off}
On the surface, the use of KV caches may seem merely a space-time tradeoff (trading KV cache storage space for shorter prefill), but the tradeoff is favorable for two reasons:
\begin{packeditemize}

\item {\em KV caches are reused a lot.}
Many contexts, especially long contexts, are frequently reused by different queries and different users. 
This can be viewed as the LLM's version of the Pareto's rule: an LLM, like human, uses 20\% of the knowledge for 80\% of the time, which means knowledge is frequently reused. Just consider that if a user asks the LLM to read a book, it is unlikely that they will only ask one book-related question.

\item {\em The size of KV caches increases slower than prefill delay.}
As the context increases, the KV cache size grows {\em linearly} whereas the prefill delay grows {\em superlinearly}.
And as the LLM gets bigger, more compute will happen at the feedforward layers which do not affect the KV cache size. 

\end{packeditemize}

\section{Knowledge Delivery Networks for Efficient Knowledge Injection}



\mypara{Limitations of Existing KV-Cache Systems}
Despite the promise, existing KV-caching learning systems still have some technical limitations. 
\begin{packeditemize}

\item {\em Limited storage for KV caches:} 
Currently, many serving engines only use the GPU/CPU memory locally accessible by an individual LLM instance to store KV caches.
Such local-memory-only storage greatly {\em limits} the amount of KV caches that are stored and reused. For instance, a CPU memory of 64 GB can only store the KV cache of 160K tokens (two pdf reports) for a small-size LLM (Llama-34B).
However, expanding the storage of KV caches to disk or remote servers would significantly constrain the bandwidth for loading the KV caches into GPU memory.

\item {\em Prefix-only reusing:}
To reuse the KV cache, most systems require that the text of the KV cache must be the {\em prefix} of the LLM input. 
Even though reusing the KV cache of the prefix is naturally supported by LLMs, this assumption of ``sharing prefix only'' severely limits its use cases.
For instance, retrieval-augmented generation (RAG) concatenates {\em multiple} retrieved text chunks in the LLM input, so most reused texts will not be the prefix. 

\item {\em Degraded quality with long contexts:}
Finally, as more texts are added to the input as LLM's context (\ie long context), the LLM's capability to capture important information might degrade, lowering the inference quality. 
Thus, when the KV caches are reused by more queries repeatedly, the degraded quality will also affect more queries. 

\end{packeditemize}

\begin{table*}[ht!]
\small
\begin{tabular}{c|c|cc}
\hline
                                                                     & \textbf{Modularity}          & \multicolumn{2}{c}{\textbf{Efficiency}}                                              \\ \hline
                                                                     & Time to inject new knowledge & \multicolumn{1}{c|}{Inference cost (\$) per request} & Response delay (s) per request \\ \hline\hline
Fine-tuning                                                          & 10 hours                     & \multicolumn{1}{c|}{0.0052}                      & 2.63                           \\ \hline
In-context learning                                                  & 0                            & \multicolumn{1}{c|}{0.0149}                      & 10.91                           \\ \hline
\rowcolor{LightCyan}KV-cache learning w/ KDN & 0.25 hours                   & \multicolumn{1}{c|}{0.0059}                      & 2.97                           \\ \hline
\end{tabular}
\vspace{14pt}
\tightcaption{Comparison between different knowledge-injection methods under a RAG setup. With KDN, KV-cache learning is $40\times$ faster when incorporating new knowledge than fine-tuning, and achieves $3.7\times$ faster and $2.5\times$ cheaper during inference compared to in-context learning.}
\vspace{-8pt}
\label{tab:comparison}
\end{table*}

\vspace{-5pt}
\subsection{Knowledge Delivery Architecture}

At first glance, these challenges facing prior KV-cache-based systems may look disparate.
Yet, our insight is that they share a common need---{\bf \em a separate KV-cache management system, which dynamically compresses, composes, and modifies KV caches to optimize the storage and delivery of KV caches and the LLM inference based on KV caches}.
We refer to such a system as a {\bf \em Knowledge Delivery Network} ({\bf KDN}).
As depicted in Figure~\ref{fig:kdn}, the envisioned architecture of a KDN consists of three main modules:
\begin{packeditemize}
    \item The {\bf \em storage} module stores the KV caches keyed by various texts and offline {\em modifies} the KV cache content such that the inference quality will be improved when the KV caches are fed to the LLM.
    \item The {\bf \em delivery} module transmits the {\em compressed} KV caches from the storage device to the server running the LLM and decompresses them in GPU to be used by the LLM serving engine.
    \item The {\bf \em blending} module dynamically {\em composes} multiple KV caches corresponding to different texts when these texts are put together in the context of an LLM query.
\end{packeditemize}

The existing LLM-serving systems~\cite{kwon2023vllm, huggingface_tgi_2024, sglang} deeply couple KV caches with LLM engines. 
In contrast, the concept of KDN is to {\bf\em separate} the management of KV caches and the LLM serving engines. 
Such separation will enable innovations on the storage, sharing, and use of KV caches, {\em without} needing to be deeply coupled with the fast-evolving LLM serving engines.
By implementing these optimizations separately, a KDN serves as a critical new system module for LLM-serving ecosystems. 
It enables the decoupling of KV-cache management from LLM serving engines, which allows LLM serving engines to focus on fast query-driven inference, while the KDN can focus on the KV-cache-related optimizations independently.

\vspace{-5pt}
\subsection{Technical Roadmap}
\label{sec:kdn}

The architecture of KDN itself does not directly address the challenges associated with KV caches; it merely breaks a potential solution into three modules. 
Fortunately, we observe that emerging research efforts could lead to a sufficient design for each module, thus making KDN practical.

\mypara{KV-cache delivery}
The recent KV cache compression techniques make it possible to cheaply store and quickly load KV caches outside GPU and CPU memory.
For instance, CacheGen~\cite{cachegen} compresses the KV cache by quantizing and then encoding it into binary strings.
LLMLingua~\cite{jiang2023llmlingua} introduces a smaller language model to identify and remove non-essential tokens in the knowledge's text, thus reducing the size of the corresponding KV cache. 
H2O~\cite{zhang2023h2o} directly removes elements in the KV cache based on their importance calculated during the inference. 
By combining the above techniques, the memory footprint of the KV cache can be reduced by over 10$\times$, drastically improving the loading speed and the storage cost of the KV cache.

\mypara{KV-cache blending}
Some recent works also improve the composability of the KV caches.
CacheBlend~\cite{yao2024cacheblend}, for instance, enables arbitrarily composing different KV caches by recomputing the cross-attention between KV caches, where the recomputation only needs 10\% computation of prefilling the full text.
PromptCache~\cite{gim2023prompt} lets users define a prompt template with different segments, which allows each segment's KV cache to be reused at different positions rather than prefix-only.

\mypara{Offline KV-cache editing}
By separating KV-cache management from the LLM-serving engines, KDNs open up new possibilities to improve inference quality.
Recent works have shown that if the attention matrix of an LLM input is appropriately modified, the inference quality can significantly improve~\cite{zhang2023tell, anonymous2024model}.
In other words, when a KV cache is ``deposited'' to KDN, KDN not only stores it but also can actively influence the LLM inference quality by offline editing the KV cache and returning the edited KV cache when it is retrieved next time, all of which is done {\em without any change to the model itself or the input prompt}.

\mypara{Interface with LLM serving engines}
Currently, most LLM serving engines, such as vLLM~\cite{kwon2023vllm}, HuggingFace TGI~\cite{huggingface_tgi_2024}, and SGLang~\cite{sglang}, do not readily support the injection of externally provided KV caches as part of the LLM input. 
Instead, their internal implementation KV-cache management (\eg paged memory in vLLM) is deeply cobbled with the model inference implementation. 
One way to deploy KDN is to augment each LLM engine with a KDN as submodule of the LLM engine. 
However, as elaborated in \S\ref{sec:kdn}, developing a performant KDN is a substantial undertaking, so it will cause much redundant engineering effort if each LLM engine maintains and evolves its KDNs. 
To avoid reinventing the wheel, the LLM serving engines could interface with a separate, shared KDN provider, via two APIs:
\textit{(i)} the LLM {\em stores} the KV cache with the associated text to KDN, and
\textit{(ii)} the LLM {\em retrieves} the KV cache from KDN using some text. 
Exposing these APIs is feasible, given most popular LLM engines are open-source.
Still, several design questions remain.
How to leverage heterogeneous links, such as NVLink, RDMA, or PCIe, to transfer KV caches?
Can the APIs be shared with other LLM functions, like disaggregated prefilling?


\mypara{Early promise}
Based on these techniques, we implemented LMCache (\url{https://github.com/LMCache/LMCache}), a prototype of KDN.
We compare the modularity and efficiency of KV-cache learning with KDN to fine-tuning and in-context learning under a RAG use case with the following setup:
\begin{packeditemize}
    \item Total size of knowledge (in text): 2 million tokens.
    \item Each request: 8K tokens knowledge plus 2K tokens user chatting history\footnote{We assume the chatting history cannot be pre-learned by fine-tuning.}.
    \item Language model: Llama 3.1 70B~\cite{dubey2024llama}.
    \item Hardware: 2 Nvidia A40 GPUs.
\end{packeditemize}
Modularity is measured by the time\footnote{The time for fine-tuning the model is estimated from Llama-Adapter~\cite{zhang2023llamaadapter}.} of injecting the knowledge base into LLM, and efficiency is measured by inference cost\footnote{The inference cost is calculated based on the AWS cloud price~\cite{awsec2pricing}.} and response delay per request.
Table~\ref{tab:comparison} shows that with the KDN, KV-cache learning can be 40$\times$ faster than fine-tuning when injecting new knowledge, and it also achieves $3.7\times$ cheaper and 2.5$\times$ faster during inference time compared to in-context learning.





\section{Conclusion}
In short, this paper makes a case for 
(1) the separation between the management of knowledge, in the form of KV caches, and LLM serving engines, and 
(2) a Knowledge-Delivery Network (KDN) as a new LLM system component that harnesses recent research development to optimize the efficiency (in speed and cost) of KV-cache-based knowledge injection.
We hope this paper can inspire more research to tackle the aforementioned problems.

\bibliographystyle{plain}
\bibliography{reference}

%

\end{document}